\documentclass{article}

\usepackage[final]{neurips_2023}

\usepackage{algorithm}
\usepackage{algpseudocode}

\usepackage{graphicx}
\usepackage[utf8]{inputenc} %
\usepackage[T1]{fontenc}    %
\usepackage{hyperref}       %
\usepackage{url}            %
\usepackage{booktabs}       %
\usepackage{amsfonts}       %
\usepackage{nicefrac}       %
\usepackage{microtype}      %
\usepackage{graphicx}
\usepackage{listings}
\usepackage{enumitem}

\usepackage{fontawesome}
\usepackage{tabularx}
\usepackage[table]{xcolor}  %
\algnewcommand{\LeftComment}[1]{\Statex \hspace{0.4cm} \(\triangleright\) #1}
\usepackage{array} 
\newcolumntype{L}[1]{>{\raggedright\arraybackslash}p{#1}} %

\usepackage[capitalize]{cleveref}
\Crefname{proposition}{Prop.}{Props.}
\Crefname{section}{Sec.}{Secs.}
\Crefname{table}{Tab.}{Tabs.}
\Crefname{appendix}{App.}{Apps.}
\usepackage{makecell}

\newcommand{\NoIndentLeftComment}[1]{\noindent\(\triangleright\) #1}

\usepackage[subtle,mathspacing=normal]{savetrees}
\usepackage[light,scaled=0.85]{roboto-mono}         %
\usepackage{caption}                                %
\newcommand{\paragraphtight}[1]{\par\textbf{#1}~~~}

\title{Capture the Flag: Uncovering Data Insights with Large Language Models}

\author{%
  Issam Laradji, Perouz Taslakian, Sai Rajeswar, Valentina Zantedeschi, Alexandre Lacoste,\\
  \textbf{Nicolas Chapados, David Vazquez, Christopher Pal, Alexandre Drouin} \\
  ServiceNow Research\\
  Montréal, Canada\\
  \texttt{firstname.lastname@servicenow.com} \\
}

\begin{document}

\maketitle

\begin{abstract}
The extraction of a small number of relevant insights from vast amounts of data is a crucial component of data-driven decision-making.
However, accomplishing this task requires considerable technical skills, domain expertise, and human labor.
This study explores the potential of using Large Language Models (LLMs) to automate the discovery of insights in data, leveraging recent advances in reasoning and code generation techniques.
We propose a new evaluation methodology based on a ``capture the flag'' principle, measuring the ability of such models to recognize meaningful and pertinent information (flags) in a dataset.
We further propose two proof-of-concept agents, with  different inner workings, and compare their ability to capture such flags in a real-world sales dataset.
While the work reported here is preliminary, our results are sufficiently interesting to mandate future exploration by the community.
\end{abstract}

\section{Introduction}

Many organizations---businesses, government agencies, and academic research groups---accumulate vast amounts of diverse data with the intent of using it for decision-making~\citep{mcafee2012bigdata, colson2019aidriven,bean2022datadriven}. Accumulated historical data can empower organizations to acquire valuable insights, make informed decisions, and be used to make predictions for future business scenarios and their respective likelihoods~\citep{colson2019aidriven}. Such insights can allow more effective interventions in tasks that are typically conducted based on intuition~\citep{mcafee2012bigdata}. However, the true value of this data only becomes apparent when individuals possess the necessary resources, time, and expertise to fully leverage it~\citep{bean2022datadriven}. Extracting meaningful visualizations, generating insightful summaries, and identifying anomalies all require a significant level of skill~\citep{mcafee2012bigdata}. Moreover, interpreting these insights effectively extends beyond data manipulation, calling for extensive domain expertise~\citep{colson2019aidriven}.

In this work, we envision autonomous data-science agents capable of extracting insights and interpreting them within context; such agents would enable individuals with low data-science expertise (e.g., business decision-makers) to make the most out of their data.
We further seek agents capable of surfacing insights of key interest, such as to limit the cognitive burden of decision-makers and mitigate the effect of cognitive biases~\citep{colson2019aidriven}.
We argue that approaches that rely solely on statistical analysis, such as the \emph{Automatic Statistician}~\citep{Steinruecken2019}, are of limited interest for this task since they lack the background knowledge required to interpret their findings in context.
Rather, we hypothesize that Large-Language models (LLMs), such as ChatGPT~\citep{chatgpt, shen2023hugginggpt} and GPT-4~\citep{OpenAI2023GPT4TR}, may serve as a good basis for such agents in light of recent advances in reasoning~\citep{wei2022chain}, code generation~\citep{olausson2023demystifying}, and given the breadth of their knowledge bases~\citep{bubeck2023sparks}.
This direction has been increasingly explored by the community, with some success, e.g., LiDA~\citep{dibia-2023-lida}, GPT4-Analyst~\citep{cheng2023gpt4}), and Sheet-Copilot~\citep{li2023sheetcopilot}.

To facilitate the development of such agents, we propose a new evaluation methodology, based on a \emph{capture the flag} principle.
In contrast with the common approach, which consists in evaluating the correctness of code generated by agents~\citep{huang-etal-2022-execution, yin-etal-2023-natural, shinn2023reflexion}, we focus on the end result and assess the ability of agents to recover key insights, regardless of how they were obtained.
This results in a general approach that can be used to evaluate all kinds of agents, irrespective of their internal mechanisms.
Our contributions are as follows.

\paragraphtight{Contributions:}
\begin{itemize}[leftmargin=4mm]
    \item We propose a \emph{capture the flag} approach to evaluating data science agents (\cref{sec:capture})
    \item We elaborate two LLM-based proof-of-concept agents for this task (\cref{sec:agents})
    \item We assess the ability of such agents to capture flags planted in real-world sales data (\cref{sec:results})
\end{itemize}
We wish to highlight that this work serves as a proof-of-concept, intended to inspire further developments in the community, rather than providing a comprehensive methodology and benchmark.

\begin{figure}
    \centering
    \includegraphics[width=\linewidth]{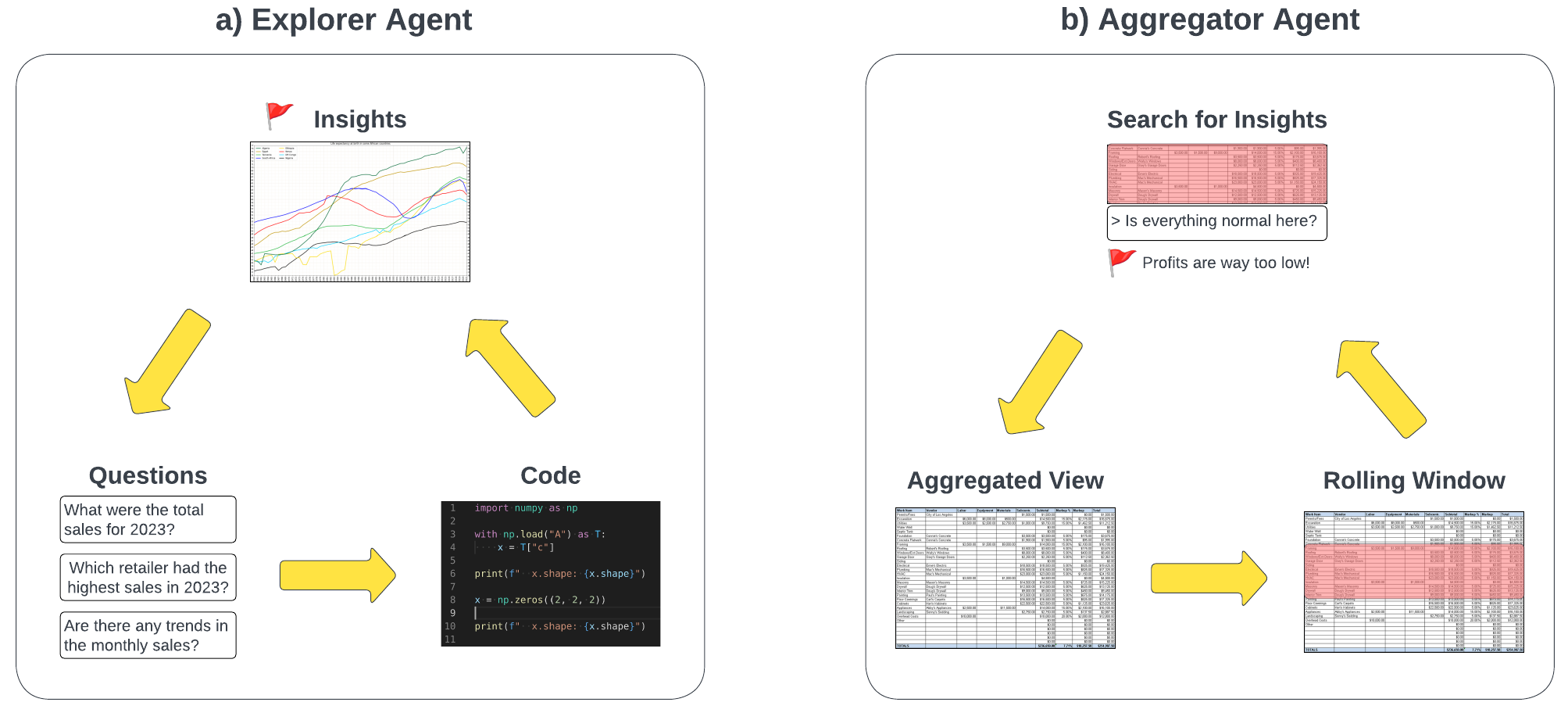}
    \caption{Overview of our data science agents. a) The Explorer agent, which generates questions and writes code to answer them; b) The Aggregator agent, which produces various aggregations of the data and then looks at snippets of the aggregated data, pointing out anything it finds relevant. Both agents can perform multiple cycles, using the discovered insights to guide future exploration.}
    \label{fig:agent-overview}
\end{figure}

\section{Related Work}

\paragraphtight{Automatic Data Analysis:}
Considering the magnitude of skills and expertise needed in data science, it is essential to develop automated systems for this process~\citep{Aggarwal2019HowCA, analyza17, Raedt2018ElementsOA, Dibia2018Data2VisAG}. In their work \citet{Steinruecken2019, Meduri2021BIRECGD} make an effort to streamline multiple aspects of data science, including the automated creation of models from data and the generation of reports with minimal human involvement. These reports not only incorporate fundamental plots and statistics, but also feature human-friendly explanations presented in natural language.
Recent progress in AI suggests that LLM-based analysis assistants are a promising solution, enabling data analysts to execute and automate their analyses~\citep{Chandel2022TrainingAE, yin-etal-2023-natural}. \citet{dibia-2023-lida, maddigan2023chat2vis} investigate GPT's capabilities in the realm of visualization tasks, primarily concentrating on its proficiency in code generation for data visualizations. \citet{gu2023analysts} provides design guidelines to empower end-user analysts interacting with AI-based analysis assistants.

\paragraphtight{Code Generation:}
LLMs can generate highly expressive programs in response to natural language prompts. These programs are proficient in addressing a wide range of tasks spanning competitive programming~\citep{Li2022CompetitionlevelCG}, data science~\citep{Lai2022DS1000, huang-etal-2022-execution, Wang2022ExecutionBasedEF}, all guided by high-level descriptions. Code generation models based on LLMs~\citep{Chen2021EvaluatingLL, Khan2023, Fried2022InCoderAG, Agashe2019JuICeAL} alongside program synthesis techniques~\citep{Chaudhuri2021NeurosymbolicP, Zhang2021InterpretablePS} empower users to accomplish tasks without the need for traditional programming approaches. Code generation for data analytics have seen a rise in research efforts, tackling important issues in automating the creation of data analysis pipelines. 
\cite{Lai2022DS1000, huang-etal-2022-execution} propose a benchmark in the Data Science domain that can be evaluated using human-written test cases of diversified Python libraries. \citet{yin-etal-2023-natural} investigate Python Jupyter Notebooks as the preferred platform for interactive coding, while limiting permissible actions to a restricted scope of code and libraries. \citet{chen-etal-2021-plotcoder} generate visualization code from natural language but limit options to a subset of Matplotlib.
In this work, we deviate from the code-generation setting as we attribute more focus to the solution, rather than how it was obtained.

\paragraphtight{GPT for Data Analysis:}
GPT models have the potency to proficiently analyze, and extract valuable insights from extensive unstructured text datasets~\citep{autogpt}. \citet{chen2023generating} examine GPT models (GPT-3.5 and GPT-4) in the context of a data visualization course, shedding light on their advantages and drawbacks.
\citet{ma2023} introduce the InsightPilot system, an automated data exploration tool leveraging LLMs to streamline the data exploration process. It achieves this by automatically identifying suitable analysis objectives and generating corresponding targeted queries, known as intentional queries (IQueries). \citet{cheng2023gpt4, wang2023plan} evaluate GPT-4's data analysis abilities and introduces an end-to-end automation framework for data processing.

\paragraphtight{Anomaly Detection:}
The problem of anomaly detection has been an active area of study with important applications in many industries. 
Statistical approaches for finding outliers and, more recently, deep learning methods have shown great success in addressing the problem~\citep{salehi2022a, ruff21, hendrycks2017a, yang2022openood}. However, anomaly detection remains a challenging task, as the definition of what constitutes an anomaly is vague: it often depends on the industry and requires specialized background to detect which of the outliers are in fact anomalous~\citep{Foorthuis2020OnTN}. However, the rarity of anomalies in real-world scenarios poses a challenge for machine learning training due to the lack of labeled data samples~\citep{NEURIPS2020_shen, NEURIPS2019_7895fc13}.
In this work, we leverage the background knowledge of LLMs to help identify anomalies.

\section{Capturing the Flag as an Evaluation Methodology}
\label{sec:capture}
We propose to evaluate data science agents based on their ability to recover salient facts in data.
Our evaluation disregards the methodology utilized by the agents in discovering such facts, thereby enabling the assessment of agents with varying inner workings.
Our approach is straightforward. We select an arbitrary dataset and manually introduce corruption to incorporate deviations from what could be expected in the context of the dataset. We term such deviations \emph{flags}.
We then evaluate the ability of agents to recover the planted flags. This approach is inspired by the one used in cybersecurity competitions~\footnote{See example competition here: \url{https://infosec-conferences.com/event-series/def-con/}}, where participants are challenged to detect and recover flags representing vulnerabilities or malicious patterns in software systems.

\paragraphtight{What Makes a Good Flag?} A good flag is an observation that could be useful for decision-making. Therefore, flags that correspond to insignificant data corruption, such as invalid dates like Feb. 31st, are not considered. Instead, we favor flags that: (i) have a plausible contextual explanation, and (ii) necessitate some background knowledge to identify. For example, in a clothing sales transactions dataset, a flag could be created by inflating swimwear sales in January beyond their June levels. This flag is plausible, as it could result from a marketing campaign, but domain knowledge tells us that most people purchase swimsuits during the summer, not the winter.

\paragraphtight{Adidas Sales Dataset:}
Throughout the rest of this work, we use the \emph{Adidas Sales dataset}~\citep{chaudhari2023adidassales} as a running example.
This dataset contains $14$ attributes and $9652$ entries corresponding to daily information on the sales of various Adidas products, including the number of units sold, the total sales amounts, profit margins, etc.
We manually incorporate the following flags in the data and assess the ability of our proposed LLM-based data science agents to recover them:
\begin{enumerate}
    \item[\faFlag~1.] The profit margins for all retailers in the state of Arizona are extremely low ($0.1\%$)
    \item[\faFlag~2.] The state of Alaska has higher total sales than California despite its smaller population
    \item[\faFlag~3.] On one given day, one retailer in Los Angeles sold an enormous quantity of men's footwear
\end{enumerate}
These flags are interesting because they require various levels of background knowledge, as well as various levels of data aggregation to be detectable. For instance, detecting Flag (1) requires knowing that a store should normally make more than $0.1\%$ profit; Flag (2) requires knowledge of the population of various states and can only be detected when aggregating sales data at the state level; and Flag (3) is a localized event that requires a fine-grained look at the specific data sample.

\paragraphtight{Towards a Benchmark:} This preliminary work aims to assess the feasibility of creating a large-scale benchmark based on the capture-the-flag methodology. As such, we focus on the Adidas Sales dataset and defer a generalization to other datasets for future research. Additionally, we plan to address extensions, such as elaborating a methodology to account for flags that could naturally arise in such datasets in our future work.

\section{Data Science Agents}\label{sec:agents}

As a proof of concept, we propose two data science agents, with  different inner workings, and assess their ability to recover flags planted in data. The prompts used by our agents are resolved by OpenAI API calls to models such as GPT-3.5~\citep{chatgpt}. An overview of these agents is provided in \cref{fig:agent-overview} and their inner workings are detailed below.

\subsection{Explorer Agent}

This first agent relies on code generation.
It follows a \emph{top-down} approach: analyzing the schema of a dataset and generating a few questions, which it answers through code generation.
Then, it generates more questions, digging into anything that it deems interesting or surprising, and the process is repeated.
After multiple rounds of questions, it reports any noteworthy insights.
A detailed pseudocode is available in \cref{alg:explore-agent} and the prompts used at each step are given in \cref{app:explorer-agent}.

\paragraphtight{Limitations:} The main limitation of Explorer Agent is that it must ask a series of high-level questions in order to identify traces of the flags and then, in subsequent rounds of questions, dig until it fully uncovers a flag. When flags exist at a high level, e.g., Flag (2) in \cref{sec:capture}, this is not a problem. However, flags corresponding to anomalous events that occur on one given day can easily be missed if the wrong questions are asked, e.g., Flag (3) in \cref{sec:capture}.

\begin{algorithm}
\footnotesize
\scalebox{0.8}{%
\begin{minipage}{1\linewidth}
\caption{ExplorerAgent}
\label{alg:explore-agent}
\begin{algorithmic}[1] %
\Require rawData, generalGoal, nRounds, dataContext
\Ensure Top insights about the dataset

\State \textbf{Initialize:} questions $\leftarrow \{\}$, answers $\leftarrow \{\}$, insights $\leftarrow \{\}$

\While{not Reached(nRounds)}

    \LeftComment{ Stage 1: Generate questions using the goal and the accumulated insights}
    \State questions $\leftarrow$ AskLLM(rawData, generalGoal, insights, dataContext) as given by the prompt at~\cref{app:exp-questions} 
    
    \LeftComment{Stage 2: Generate code for each question}
    \For{each question in questions}
        \State code $\leftarrow$ GenerateCodeToAnswer(question, dataContext) 
        \State answer $\leftarrow$ ExecuteCode(data, code)
        \State answers.add(answer)
    \EndFor
    
    \LeftComment{Stage 3: Extract insights from answers}
    \State insights.add$\left(\right.$ExtractInsights(answers)$\left.\right)$
\EndWhile

\State \textbf{Return} TopInsights(insights) as given by the prompt at~\cref{app:exp-extract} 

\end{algorithmic}
\end{minipage}
}
\end{algorithm}

\subsection{Aggregator Agent}

The second agent uses a \emph{bottom-up} approach that completely differs from the previous one.
It begins by using code generation to produce various aggregations (views) of the data.
Then, it scans each of these views using a sliding window, flagging anything that appears abnormal.
Of particular interest, the flagging is done without relying on code generation and relies completely on the common-sense domain knowledge built into the model. In familiar terms, one could say that the agent \emph{eyeballs} the data.
Interestingly, in contrast to existing code-based evaluation methodologies, our capture-the-flag approach supports such agents.
A detailed pseudocode for this agent is available in \cref{alg:agg-agent} and the prompts used at each step are given in \cref{app:agg-agent}.

\paragraphtight{Limitations:} In contrast with the Explorer Agent, which is limited to asking specific questions, a key strength of the Aggregator Agent is that it looks at the data in a more holistic way. However, its main limitation is that its conclusions are not grounded in code. Rather, the data is added to the prompt and the agent extracts any insight it finds relevant. Proceeding in such a way makes the agent vulnerable to hallucinating facts. Further, the reliance on a sliding window approach may prevent the agent from detecting patterns that occur over multiple such windows (e.g., trends that arise over several months).

\begin{algorithm}
\footnotesize
\scalebox{0.8}{%
\begin{minipage}{1\linewidth}
\caption{AggregatorAgent}
\label{alg:agg-agent}
\begin{algorithmic}[1]
\Require {rawData, nAggregations, generalGoal, dataColumns, dataStats, slidingWindow}
\Ensure Top insights about the dataset
\State \textbf{Initialize:} aggregationList $\leftarrow \{\}$, insightCount $\leftarrow 0$, insightList $\leftarrow \{\}$

    \NoIndentLeftComment{Stage 1: Generate Aggregations of the Data}
    \While{not Reached(nAggregations)}
        \State aggregation $\leftarrow$ AskLLM(rawData, generalGoal, dataColumns, dataStats) as given by the prompt in~\cref{app:agg} 
        \State aggregatedData $\leftarrow$ AggregateData(rawData, aggregation)
        
        \State aggregationList.add(aggregatedData)
    \EndWhile

    \NoIndentLeftComment{Stage 2: Extract Insights from each aggregation}
    \For{aggregatedData in aggregationList}
     
        \For{$i$ in range(0, length(aggregatedData), slidingWindow)}
            \State aggregatedDataWindow $\leftarrow$ aggregatedData[i:i+slidingWindow]
            
            \State insights $\leftarrow$  ExtractInsights(aggregatedDataWindow) as in~\cref{app:extract} 
            \State insightList.add(insights)
              
        \EndFor
    \EndFor
      \State \textbf{Return}  TopInsights(insights)  using the prompt in~\cref{app:rank}
\end{algorithmic}
\end{minipage}
}
\end{algorithm}

\begin{table}[ht]
\centering
\caption{Insights extracted by the Aggregator agent.
The table shows all flag-related insights that were discovered as well as a representative sample of other, non-flag-related, insights.
Each insight is accompanied by a description of the aggregated view in which it was found, the numerical value of the quantity in question, a generated explanation of its relevance, and how the data was aggregated in terms of which column it was grouped by and the column it was aggregated over.}
\renewcommand{\arraystretch}{1.15}
\footnotesize
\resizebox{\textwidth}{!}{%
    \begin{tabular}{@{}L{4cm}p{3cm}lp{5.3cm}@{}}
    \toprule
    \textbf{Insight} & \textbf{Aggregation} &  \textbf{Value} &\textbf{Explanation} \\
    \toprule
    \multicolumn{4}{c}{\faFlag~~\bf{Flag 1}} \\
    \midrule
    Arizona has an extremely low Operating Margin & \makecell[l]{Grouped by: \bf{State}\\
    on \bf{Operating Margin}} & 0.1 \%  & This insight is captivating because it demonstrates an extremely low operating margin, indicating very low profitability for this sale. \\
    \midrule
    \multicolumn{4}{c}{\faFlag~~\bf{Flag 2}} \\ 
    \midrule
    Alaska has the highest sales. & \makecell[l]{Grouped by: \bf{State}\\
    on \bf{Total Sales}} & \$ 49\,473\,404 & Despite its small population, Alaska has the highest sales, which is surprising considering the larger populations and economies of states like California and New York.\\
    \midrule
    \multicolumn{4}{c}{\faFlag~~\bf{Flag 3}} \\ 
    \midrule
    Kohl's has unusually high units sold. & None & 8\,000\,000 & This is the highest number of units sold, which is surprising given the average units sold in other rows.\\
    \midrule
    \multicolumn{4}{c}{\bf{Other}} \\ 
    \midrule
    West Gear, a lesser-known retailer, has high sales. & \makecell[l]{Grouped by: \bf{Retailer}\\
    on \bf{Units Sold}} & 715\,900 & It's surprising that a lesser-known retailer like West Gear has such high total sales, nearly matching Amazon.\\
    \midrule
    Kohl's has highest Units Sold & None & 7\,500 & This insight is the most interesting as it highlights the highest number of units sold, indicating a very popular product. \\
    \midrule
    Men's Street Footwear has the highest Total Sales & None & 732\,000~\$ & This insight is intriguing because it represents the highest total sales value, indicating a very successful sale. \\
    \midrule
    Foot Locker has the highest Operating Margin seen & None &  65 \% & This insight is interesting as it reveals the highest operating margin in the data, suggesting a very profitable sale. \\
    \bottomrule
    \end{tabular}
}
\label{tab:generated-insights-agg}
\end{table}

\begin{table}[ht]
\centering
\caption{Insights extracted by the Explorer agent. Each insight is accompanied by the question that led to its discovery and a generated explanation of its relevance. Insights corresponding to planted flags are shown first, followed by other non-flag-related insights.}
\renewcommand{\arraystretch}{1.15}
\footnotesize
\resizebox{\textwidth}{!}{%
    \begin{tabular}{@{}L{5cm}p{5cm}p{5.4cm}@{}}
    \toprule
   \textbf{Question} & \textbf{Insight} & \textbf{Explanation} \\
    \toprule
    \multicolumn{3}{c}{\faFlag~~\bf{Flag 1}} \\
    \midrule
    What is the relationship between operating margin and sales method for each retailer in 2021?
 & Kohl's had a very low average operating margin for online sales (0.001\%), but a higher average operating margin for outlet sales (35.63\%). & The almost non-existent operating margin for Kohl's online sales, when juxtaposed with its outlet sales, is surprising. Retailers typically achieve significant margins through online sales due to reduced overheads compared to physical stores.\\
    \midrule
    \multicolumn{3}{c}{\faFlag~~\bf{Flag 2}} \\ 
    \midrule
What are the top 5 cities in terms of sales revenue for 2021?
& Anchorage is the top city in terms of sales revenue for 2021.   & Anchorage being at the top in sales revenue is unexpected, given that it's a less populated city when compared to other major U.S. cities. One would expect cities with higher populations and larger economies, like San Francisco or Chicago, to lead in sales revenue. \\
    \midrule
    \multicolumn{3}{c}{\faFlag~~\bf{Flag 3}} \\ 
    \midrule
    \multicolumn{3}{c}{\textit{Agent failed to capture the flag}} \\
    \midrule
    \multicolumn{3}{c}{\bf{Other}} \\ 
    \midrule
    What is the average units sold per transaction for each sales method (Outlet, Online, In-store) in 2021?
 & In-store and Outlet sales methods have a significantly higher average units sold per transaction compared to Online. & In today's digital age, with the rise of e-commerce platforms, one might expect online sales to have higher transactions. However, the data suggests that physical outlets still have robust sales.\\
    \midrule
     Are there any seasonal trends in sales for specific product categories in 2021?
 & All product categories have the highest sales in January. & While some January sales might be expected due to post-holiday clearances, it's surprising to see all categories peak during this month, suggesting a broad trend or seasonality.\\
    \midrule
         How did the sales of each retailer evolve month by month in 2021? &
         Sports Direct had no sales data available for any month in 2021.
 &  While not shocking, it is noteworthy that there's no data for a retailer for the entire year. This could indicate a data issue or that the retailer was not operational. \\
    \midrule
         What is the relationship between operating margin and total sales for each product category in 2021?
 & There is no strong correlation between operating margin and total sales for product categories in 2021.& A business might aim for a balance between sales volume and profit margins. The absence of a strong correlation indicates that some categories might be focusing on volume, while others prioritize profitability.\\
    \bottomrule
    \end{tabular}
}
\label{tab:generated-insights-explorer}
\end{table}

\section{Experiments}\label{sec:results}

In this section, we assess the ability of our agents to capture flags planted in the Adidas Sales dataset.
We start by reviewing the experimental protocol and then discuss the results separately for each agent.

\subsection{Protocol}

Our experiments focus on the three flags presented in \cref{sec:capture}.
For each flag, we create a separate copy of the original Adidas Sales dataset and apply the corruption that corresponds to planting the flag.
Given the potential cost associated with OpenAI API calls, we have opted to limit our datasets to 1000 rows, by selecting 100 rows for each of the following states: New York, Texas, California, Illinois, Arizona, Alaska, Colorado, Washington, Florida, and Minnesota.
We then run each agent on these datasets and instruct them to rank their insights according to relevance using the prompts shown in \cref{app:agents}.

The extracted insights are then evaluated in terms of whether i) they correspond to planted flags, ii) the data is factual (not hallucinated), and iii) the insight is relevant. In what follows, we review and discuss the results for each agent.

\subsection{Aggregator Agent Results}

\paragraphtight{Implementation Details.}
We employed the GPT-3.5 model with a sliding window of size 50 to process the input data, generating 20 aggregations for the data and extracting 5 insights from each window. Subsequently, we utilized the GPT-4 model to rank and refine the final set of insights based on how interesting they are to a data analyst as shown in~\cref{alg:agg-agent}. See ~\cref{app:agg-agent} for an explanation of the prompts and the variables used in the Algorithm.

\paragraphtight{Relevance.}  Table~\ref{tab:generated-insights-agg} presents insights generated by the Aggregator agent, with each insight accompanied by an explanation of its significance. Interestingly, all flags were identified among the top 5 generated insights.
For example, for Flag 2, the agent mentions that it was reported since it is surprising that Alaska has the highest sales despite its small population in comparison to California and New York.
It is worth noting that the agent also returns insights that are not part of the planted flags, but that appear to be relevant.
For example, it reports that the sales amount of retailer West Gear nearly match those of Amazon, which is surprising given that the former is lesser-known.
Finally, we observe that the reported insights were extracted from both aggregated data and the original data, supporting the relevance of considering both kinds of data in the analysis.

\paragraphtight{Consistency.}  We have learned that when we prompt the agent to be specific about which columns, values, and row numbers it bases its insights on (as demonstrated in the Prompts section in the Appendix~\ref{app:agg-agent}), the results can be easily verified after they are generated. We can automatically cross-check the results against our dataset programmatically by asserting that the values at the cited row and columns match.  As a result, the insights we get (as shown in Table~\ref{tab:generated-insights-agg}) are always correct in terms of the reported data values. However, some non-numerical insights, like ``Kohl's has the highest number of sales'' are more difficult to validate and may not hold true in the data. Therefore, we plan to investigate LLM-based methods that can create code that specifically verifies if these claims are true in future work.

\paragraphtight{Limitations.} There are a couple of important limitations to this agent. First, using the window approach with large amounts of data leads to many LLM queries, which can be quite expensive. Hence, we opted for using GPT-3.5 instead of the better, but more expensive, GPT-4. This means that our current agent may not be using this suite of models to its full potential. Second, our approach tends to provide insights that are local in view, making it difficult to see big trends or broader patterns in the data. In other words, it is challenging to get a high-level view with our bottom-up method which can potentially be addressed by the Explorer Agent.

\paragraphtight{Future Work.} While our research has shown promising initial results in extracting useful insights from data, there is room for improvement in achieving deeper insights. We intend to experiment with other language models like GPT-4~\citep{OpenAI2023GPT4TR} and Llama-2~\citep{touvron2023llama}, as well as explore different ways to prompt the agent to perform data aggregation. These explorations could lead to methods that are crucial in extracting more interesting insights. We currently prompt the agent to use first- and second-order dataset statistics for aggregating the data, but, in some cases, higher statistics could be necessary to identify interesting insights. Another important aspect is to investigate other forms of aggregations like computing correlations between columns instead of grouping them based on different first-level statistics. Further, there are many hyperparameters that need to be explored, like how many aggregations we need, the sliding window size, and the number of insights to be extracted from each portion of the data. We intend to address these limitations in future work; however, our current work serves as a thought-provoking proof of concept, offering valuable ideas and discussion points.

\subsection{Explorer Agent Results}~\label{sec:explorer}

\paragraphtight{Implementation Details.} We employed the GPT-3.5 model for generating the outputs for our prompts and we have set the number of rounds of question refinement to be 3, defined as nRounds in Algorithm~\ref{alg:explore-agent}. We defined the high level goal generalGoal to be  ``I want a general overview of the sales for 2021.''  and the dataContext to be ``This is a dataset of sales transactions'' which we both set in the prompt to help in generating the questions and the code (prompt can be found here~\ref{app:explorer-agent}). We ask the model to generate 10 questions for the data in each round. The dataSchema is the description of each column in the rawData which is a csv file where each row represents a record corresponding to the data. 

\paragraphtight{Relevance.} 
Table~\ref{tab:generated-insights-explorer} presents insights extracted by the Explorer agent, categorized into flagged and non-flagged insights. These insights are highly relevant to the user as they address specific questions, such as the relationship between operating margin and sales method, top-performing cities in terms of sales revenue, and seasonal trends in sales for product categories in 2021. Flagged insights (Flag 1 and Flag 2) highlight unexpected findings, such as Kohl's low operating margin for online sales and Anchorage (Alaska) leading to high sales revenue despite its smaller population. The non-flagged insights also provide valuable information, such as the higher average units sold per transaction in physical stores and outlets compared to online sales and the lack of sales data for Sports Direct throughout 2021. Additionally, the absence of a strong correlation between operating margin and total sales for product categories suggests varying business strategies among retailers. These insights collectively offer valuable information for informed decision-making and analysis.

\paragraphtight{Consistency.}  All values extracted by the Explorer Agent in Table~\ref{tab:generated-insights-explorer} have been derived directly from the output of the code generated to address each specific question, ensuring their factual basis. However, it is important to note that, similar to the Aggregator Agent, the justifications themselves may require further verification by a critic (which could be human or an efficient LLM method). These justifications are judgments made by the model and should be subject to scrutiny and validation to ensure their accuracy and reliability.

\paragraphtight{Limitations.} The Explorer Agent has certain limitations in its approach. It begins by asking high-level questions to identify patterns in the data, which may inadvertently overlook crucial flags that would be interesting to the data analyst. For example, Flag 3, which consists of an event that happened in a very short period, is overlooked by this agent. This likely occurs due to its tendency to ask high-level questions, rather than digging into low-level details. Furthermore, it does not consider the local perspectives of the data, potentially missing deviations from expected norms, an aspect that the Aggregator Agent could potentially address. Another limitation is that the Explorer Agent might generate code that is not executable (due to bugs for example). Addressing this limitation requires additional OpenAI API calls showing the original prompt along with the execution error  in order to re-generate the code with the bugs fixed. However, generating executable code is not guaranteed, forcing the agent to skip the question altogether. These limitations highlight the need for a more comprehensive and context-aware approach to data analysis and code generation.

\paragraphtight{Future Work.} We plan to conduct more experiments related to the Explorer Agent to investigate how it behaves when using different large language models, like GPT 4, StarCoder~\cite{li2023starcoder} and Llama 2~\cite{touvron2023llama}, and when generating the questions and code based on different prompts. We plan to evaluate how changing hyperparameters like the number of questions, and the number of times we refine the questions impacts the result. We also plan to compare between different context and high level goal that we set in the prompts with respect to the quality of the generated insights. 

\section{Conclusion}
In conclusion, we explore two distinct approaches for automating data analysis and extracting useful insights from a ``capture the flag''  benchmark that we have introduced: the top-down approach of the Explorer Agent and the holistic, bottom-up, perspective of the Aggregator Agent.  The Explorer Agent excels in generating context-aware high-level questions and code, but may overlook specific flags, miss local deviations, and can potentially generate non-functional code. Meanwhile, the Aggregator Agent offers a comprehensive data view, but could potentially struggle with capturing long-term trends, and can be expensive with large datasets. Given these observations, it becomes evident that a hybrid approach, combining the high-level question generation capabilities of the Explorer Agent with the holistic data perspective of the Aggregator Agent, holds great promise for future research and development in automating data analysis effectively. This work represents a preliminary step towards effective data analysis automation using large language models, emphasizing the need for continued research in this avenue.

\bibliographystyle{plainnat}
\bibliography{references}

\appendix

\newpage
\section{Supplementary Materials}\label{app:agents}

\subsection{Explorer Agent}\label{app:explorer-agent}

\subsubsection{Generating the Questions}\label{app:exp-questions}

\begin{lstlisting}[frame=single, basicstyle=\ttfamily\footnotesize]
 Hi, I require the services of your team to help me reach my goal.

        <context>{dataContext}</context>

        <goal>{generalGoal}</goal>

        <schema>{dataSchema}</schema>

        <insights>{insights}</insights>

        Instructions:
        * Produce a list of questions to be solved by the data scientists in your
        team to explore my data and reach my goal.
        * Explore diverse aspects of the data, and ask questions that are relevant to
        my goal.
        * You must ask the right questions to surface anything interesting (trends,
        anomalies, etc.)
        * Make sure these can realistically be answered based on the data schema.
        * The insights that your team will extract will be used to generate a report.
        * Each question that you produce must be enclosed in <question></question> tags.
        * Do not number the questions.
        * You can produce at most {max_questions} questions.
\end{lstlisting}

\subsubsection{Ranking the Insights Prompt}\label{app:exp-extract}
\begin{lstlisting}[frame=single, basicstyle=\ttfamily\footnotesize]
Rank the answers and justification from the sales csv below based on the order of
how surprising each is. Start with the most surprising. Explain why these insights 
deviate from what is expected.

Write the row number, the insight, the values, and an explanation

Put it in this format.

Row:
Insight:
Explanation:

{insights}
\end{lstlisting}

where {insightList} looks like the raw CSV version of Table~\ref{tab:generated-insights-explorer}.

\subsection{Aggregator Agent}\label{app:agg-agent}

\subsubsection{Data Aggregation Prompt}
\label{app:agg}
Below is the prompt used for identifying different aggregations of the data where the values under ``CSV Columns'' and ``Stats'' are data dependent.
\begin{lstlisting}[frame=single, basicstyle=\ttfamily\footnotesize]
{generalGoal}
Below is an instruction that describes a task. Write a response that appropriately
completes the request.

### Instruction:

Given these csv columns and their stats, what are 20 useful aggregations to the data
that groups on one column and aggregates values on another column? 
Write them in this format

Groupby:
Target column:
Aggregation function:
        

CSV Columns:
=======
{dataColumns}

        
Stats
=====
{dataStats}

\end{lstlisting}
where for the sales dataset, \{generalGoal\} is
\begin{lstlisting}[frame=single, basicstyle=\ttfamily\footnotesize]
You are a sales expert analyst who is interested in understanding the operations of
the store sales across the USA.
\end{lstlisting}

\{dataColumns\} is
\begin{lstlisting}[frame=single, basicstyle=\ttfamily\footnotesize]
Retailer,Retailer ID,Invoice Date,Region,State,City,Product,Price per Unit,
Units Sold,Total Sales,Operating Profit,Operating Margin,Sales Method
\end{lstlisting}

\{dataStats\} is
\begin{lstlisting}[frame=single, basicstyle=\ttfamily\footnotesize]
Retailer ID,Price per Unit,Units Sold,Total Sales,Operating Profit,Operating Margin
count,1000.0,1000.0,1000.0,1000.0,1000.0,1000.0
mean,1163877.567,53.096,13604.612,654966.201,85163.373,35.83
std,29580.41380228183,12.881331946375449,284477.0482445489,12800683.818093507,\
71638.05466051835,9.748237180543368
min,1128299.0,20.0,7.0,224.0,112.0,10.0
25%
50%
75%
max,1197831.0,110.0,9000000.0,405000000.0,390000.0,76.0

\end{lstlisting}
\subsubsection{Extracting Insights Prompt}
\label{app:extract}
Below is the prompt used for extracting insights from a chunk of the data represented under ``CSV Data.''
\begin{lstlisting}[frame=single, basicstyle=\ttfamily\footnotesize]
{generalGoal}
Below is an instruction that describes a task. Write a response that appropriately
completes the request.

### Instruction:

Find 5 surprising, interesting insights from the csv below in 8 words max in bullet
points. For each cite the row number, explain why, provide the relevant value as
(column, value), and give a score 1-5 about how surprising it is and why did you 
give it that score.

        Row:
        Insight:
        Values:
        Score:
        Explanation:
        
CSV Data
=======
{aggregatedDataWindow}


### Response:
        
\end{lstlisting}
where an example aggregated data is \{aggregatedDataWindow\} is

\begin{lstlisting}[frame=single, basicstyle=\ttfamily\footnotesize]
,Retailer,Total Sales (sum)
0,Amazon,13158552
1,Foot Locker,64051537
2,Kohl's,417223750
3,Sports Direct,22582500
4,Walmart,38552250
5,West Gear,99397612
\end{lstlisting}

\subsubsection{Ranking the Insights Prompt}
\label{app:rank}

\begin{lstlisting}[frame=single, basicstyle=\ttfamily\footnotesize]
Rank the insights from the csv below based on order of how interesting each
is. Start with the most interesting.

Write the row number, the insight, the values,  and an explanation

Put it in this format.

Row:
Insight:
Explanation:

{CSV Content}
\end{lstlisting}

where {CSV Content} looks like the raw CSV version of Table~\ref{tab:Ainsights}.

\begin{table}[hb]
    \centering
    \caption{Insights extracted by the Aggregator Agent}
    \footnotesize
    \scalebox{0.9}{%
    \begin{tabular}{@{}L{0.25\textwidth}L{0.25\textwidth}L{0.10\textwidth}L{0.40\textwidth}@{}}
        \toprule
        \textbf{Insight} & \textbf{Values} & \textbf{Score} & \textbf{Explanation} \\
        \midrule
        Amazon's total sales are surprisingly low. & (Retailer, Amazon), (Total Sales (sum), 45,020,834) & 4 & Given Amazon's dominance in the retail market, one would expect their total sales to be higher than other retailers. \\
        \midrule
        Foot Locker has the highest total sales. & (Retailer, Foot Locker), (Total Sales (sum), 49,888,450) & 5 & It's surprising that a specialized retailer like Foot Locker has the highest total sales, surpassing even Amazon. \\
        \midrule
        Kohl's total sales are significantly lower than others. & (Retailer, Kohl's), (Total Sales (sum), 11,888,750) & 3 & Considering Kohl's is a well-known retailer, it's surprising their total sales are so much lower than other retailers. \\
        \bottomrule
    \end{tabular}
    }
\label{tab:Ainsights}
\end{table}

\end{document}